\documentclass[10pt,twocolumn,letterpaper]{article}

\usepackage[pagenumbers]{cvpr}

\usepackage{graphicx}
\usepackage{amsmath}
\usepackage{amssymb}
\usepackage{booktabs}

\usepackage{url}
\usepackage{booktabs}
\usepackage{amsfonts}
\usepackage{nicefrac}
\usepackage{microtype}
\usepackage{xcolor,colortbl}
\usepackage{graphicx}
\usepackage{tabularx}
\usepackage{tabulary}
\usepackage{multirow}
\usepackage{subcaption}
\usepackage{xfrac}
\usepackage{wrapfig}
\usepackage[pagebackref,breaklinks,colorlinks]{hyperref}
\usepackage{xspace}
\usepackage{listings}
\usepackage{upquote}

\definecolor{todocolor}{rgb}{0.82, 0.41, 0.12}

\definecolor{fixedcolor}{HTML}{333333}
\definecolor{indepcolor}{HTML}{A8165B}
\definecolor{consicolor}{HTML}{66AED2}
\definecolor{fnmatcolor}{HTML}{7ECD31}

\usepackage[capitalize]{cleveref}
\crefname{section}{Sec.}{Secs.}
\Crefname{section}{Section}{Sections}
\Crefname{table}{Table}{Tables}
\crefname{table}{Tab.}{Tabs.}

\usepackage{xcolor}

\definecolor{codegreen}{rgb}{0,0.6,0}
\definecolor{codegray}{rgb}{0.5,0.5,0.5}
\definecolor{codepurple}{rgb}{0.58,0,0.82}
\definecolor{backcolour}{rgb}{0.95,0.95,0.92}

\lstdefinestyle{mystyle}{
    backgroundcolor=\color{backcolour},   
    commentstyle=\color{codegreen},
    keywordstyle=\color{magenta},
    numberstyle=\tiny\color{codegray},
    stringstyle=\color{codepurple},
    basicstyle=\ttfamily\footnotesize,
    breakatwhitespace=false,         
    breaklines=true,                 
    captionpos=b,                    
    keepspaces=true,                 
    numbers=left,                    
    numbersep=5pt,                  
    showspaces=false,                
    showstringspaces=false,
    showtabs=false,                  
    tabsize=2
}

\def\confName{ArXiv}
\def\confYear{2022}

\newcommand{\authsep}{\;\;}

\begin{document}

\title{Better plain ViT baselines for ImageNet-1k}

\author{Lucas Beyer \authsep Xiaohua Zhai \authsep Alexander Kolesnikov \\
Google Research, Brain Team Z\"urich \\
\\
\url{https://github.com/google-research/big\_vision}
}
\maketitle

\begin{abstract}
It is commonly accepted that the Vision Transformer model requires sophisticated regularization techniques to excel at ImageNet-1k scale data. Surprisingly, we find this is not the case and standard data augmentation is sufficient. This note presents a few minor modifications to the original Vision Transformer (ViT) vanilla training setting that dramatically improve the performance of plain ViT models. Notably, 90 epochs of training surpass 76\% top-1 accuracy in under seven hours on a TPUv3-8, similar to the classic ResNet50 baseline, and 300 epochs of training reach 80\% in less than one day.
\end{abstract}

\section{Introduction}\label{sec:intro} 
The ViT paper~\cite{vit} focused solely on the aspect of large-scale pre-training, where ViT models outshine well tuned ResNet~\cite{he2016deep} (BiT~\cite{brain2020bit}) models.
The addition of results when pre-training only on ImageNet-1k was an afterthought, mostly to ablate the effect of data scale.
Nevertheless, ImageNet-1k remains a key testbed in the computer vision research and it is highly beneficial to have as simple and effective a baseline as possible.

Thus, coupled with the release of the \textit{big vision} codebase used to develop ViT~\cite{vit}, MLP-Mixer~\cite{mlp_mixer}, ViT-G~\cite{vit_g}, LiT~\cite{lit}, and a variety of other research projects, we now provide a new baseline that stays true to the original ViT's simplicity while reaching results competitive with similar approaches~\cite{deit, resnet_strikes} and concurrent~\cite{deit_iii}, which also strives for simplification.

\begin{figure}[t]
\centering
  \includegraphics[width=\linewidth]{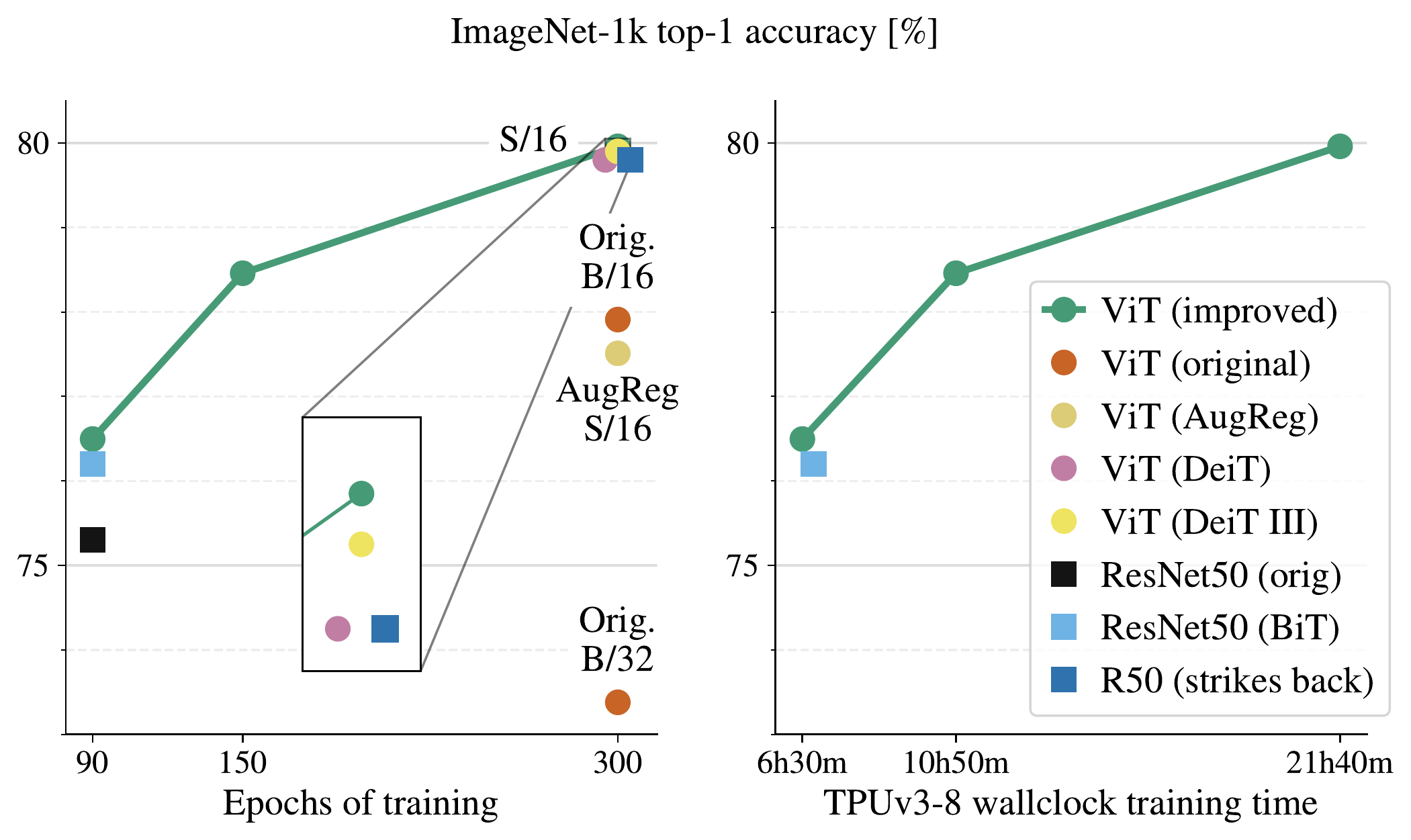}
 \caption{Comparison of ViT model for this note to state-of-the-art ViT and ResNet models. Left plot demonstrates how performance depends on the total number of epochs, while the right plot uses \texttt{TPUv3-8} wallclock time to measure compute. We observe that our simple setting is highly competitive, even to the canonical ResNet-50 setups.}
 \end{figure}

\section{Experimental setup}\label{sec:setup}

We focus entirely on the ImageNet-1k dataset (ILSVRC-2012) for both (pre)training and evaluation. 
We stick to the original ViT model architecture due to its widespread acceptance~\cite{deit, moco_v3_vit, mae, uvit, vitdet}, simplicity and scalability, and revisit only few very minor details, none of which are novel.
We choose to focus on the smaller ViT-S/16 variant introduced by~\cite{deit} as we believe it provides a good tradeoff between iteration velocity with commonly available hardware and final accuracy.
However, when more compute and data is available, we highly recommend iterating with ViT-B/32 or ViT-B/16 instead~\cite{augreg, vit_g}, and note that increasing patch-size is almost equivalent to reducing image resolution.

All experiments use ``inception crop''~\cite{inception_crop} at 224px² resolution, random horizontal flips, RandAugment~\cite{randaug},  and Mixup augmentations.
We train on the first 99\% of the training data, and keep 1\% for \emph{minival} to encourage the community to stop selecting design choices on the validation (de-facto test) set.
The full setup is shown in Appendix~A.

\section{Results}\label{sec:results}

The results for our improved setup are shown in Figure~1, along with a few related important baselines.
It is clear that a simple, standard ViT trained this way can match both the seminal ResNet50 at 90 epochs baseline, as well as more modern ResNet~\cite{resnet_strikes} and ViT~\cite{deit_iii} training setups.
Furthermore, on a small TPUv3-8 node, the 90 epoch run takes only 6h30, and one can reach 80\% accuracy in less than a day when training for 300 epochs.

The main differences from~\cite{vit, augreg} are a batch-size of 1024 instead of 4096, the use of global average-pooling (GAP) instead of a class token~\cite{moco_v3_vit,vitseecnn}, fixed 2D sin-cos position embeddings~\cite{moco_v3_vit}, and the introduction of a small amount of RandAugment~\cite{randaug} and Mixup~\cite{zhang2018mixup} (level 10 and probability 0.2 respectively, which is less than~\cite{augreg}).
These small changes lead to significantly better performance than that originally reported in~\cite{vit}.

Notably absent from this baseline are further architectural changes, regularizers such as dropout or stochastic depth~\cite{sdepth}, advanced optimization schemes such as SAM~\cite{sam}, extra augmentations such as CutMix~\cite{cutmix}, repeated augmentations~\cite{deit}, or blurring, ``tricks'' such as high-resolution fine-tuning or checkpoint averaging, as well as supervision from a strong teacher via knowledge distillation.

Table~1 shows an ablation of the various minor changes we propose. It exemplifies how a collection of almost trivial changes can accumulate to an important overall improvement.
The only change which makes no significant difference in classification accuracy is whether the classification head is a single linear layer, or an MLP with one hidden $\tanh$ layer as in the original Transformer formulation.

\begin{table}[t]
  \setlength{\tabcolsep}{0pt}
  \setlength{\extrarowheight}{5pt}
  \renewcommand{\arraystretch}{0.75}
  \newcolumntype{C}{>{\centering\arraybackslash}X}
  \newcolumntype{R}{>{\raggedleft\arraybackslash}X}
  \centering
  \vspace{-1em}
  \caption{Ablation of our trivial modifications.}
  \begin{tabularx}{\linewidth}{p{4.5cm}p{0.4cm}cp{0.4cm}cp{0.4cm}c}
    \toprule[1pt]
     && 90ep && 150ep && 300ep \\
    \midrule
    \textbf{Our improvements} && 76.5 && 78.5 && 80.0 \\
    \midrule
    no RandAug+MixUp && 73.6 && 73.7 && 73.7 \\
    Posemb: sincos2d $\rightarrow$ learned && 75.0 && 78.0 && 79.6 \\
    Batch-size: 1024 $\rightarrow$ 4096 && 74.7 && 77.3 && 78.6 \\
    Global Avgpool $\rightarrow$ [cls] token && 75.0 && 76.9 && 78.2 \\
    Head: MLP $\rightarrow$ linear && 76.7 && 78.6 && 79.8 \\
    \midrule
    \textcolor{gray}{Original + RandAug + MixUp} && \textcolor{gray}{71.6} && \textcolor{gray}{74.8} && \textcolor{gray}{76.1}  \\
    \textcolor{gray}{Original} && \textcolor{gray}{66.8} && \textcolor{gray}{67.2} && \textcolor{gray}{67.1}  \\
    \bottomrule
  \end{tabularx}
\end{table}

\begin{table}[b]
  \setlength{\tabcolsep}{0pt}
  \setlength{\extrarowheight}{5pt}
  \renewcommand{\arraystretch}{0.75}
  \newcolumntype{C}{>{\centering\arraybackslash}X}
  \newcolumntype{R}{>{\raggedleft\arraybackslash}X}
  \centering
  \vspace{-1em}
  \caption{A few more standard metrics.}
  \begin{tabularx}{\linewidth}{p{4.5cm}p{0.4cm}cp{0.4cm}cp{0.4cm}c}
    \toprule[1pt]
     && Top-1 && ReaL && v2 \\
    \midrule
    \textcolor{gray}{Original (90ep)} && \textcolor{gray}{66.8} && \textcolor{gray}{72.8} && \textcolor{gray}{52.2} \\
    Our improvements (90ep) && 76.5 && 83.1 && 64.2 \\
    Our improvements (150ep) && 78.5 && 84.5 && 66.4 \\
    Our improvements (300ep) && 80.0 && 85.4 && 68.3 \\
    \bottomrule
  \end{tabularx}
\end{table}

\section{Conclusion}\label{sec:conclusion}

It is always worth striving for simplicity. \newline

\textbf{Acknowledgements.} We thank Daniel Suo and Naman Agarwal for nudging for 90 epochs and feedback on the report, as well as the Google Brain team for a supportive research environment.

{\small
\bibliographystyle{ieee_fullname}
\bibliography{egbib}
}

\appendix
\section{{\tt big\_vision} experiment configuration}

\begin{lstlisting}[language=Python, caption=Full recommended config]
def get_config():
  config = mlc.ConfigDict()

  config.dataset = 'imagenet2012'
  config.train_split = 'train[:99%]'
  config.cache_raw = True
  config.shuffle_buffer_size = 250_000
  config.num_classes = 1000
  config.loss = 'softmax_xent'
  config.batch_size = 1024
  config.num_epochs = 90

  pp_common = (
      '|value_range(-1, 1)'
      '|onehot(1000, key="{lbl}", key_result="labels")'
      '|keep("image", "labels")'
  )
  config.pp_train = (
      'decode_jpeg_and_inception_crop(224)' +
      '|flip_lr|randaug(2,10)' +
      pp_common.format(lbl='label')
  )
  pp_eval = 'decode|resize_small(256)|central_crop(224)' + pp_common

  config.log_training_steps = 50
  config.log_eval_steps = 1000
  config.checkpoint_steps = 1000

  # Model section
  config.model_name = 'vit'
  config.model = dict(
      variant='S/16',
      rep_size=True,
      pool_type='gap',
      posemb='sincos2d',
  )

  # Optimizer section
  config.grad_clip_norm = 1.0
  config.optax_name = 'scale_by_adam'
  config.optax = dict(mu_dtype='bfloat16')
  config.lr = 0.001
  config.wd = 0.0001
  config.schedule = dict(warmup_steps=10_000, decay_type='cosine')
  config.mixup = dict(p=0.2, fold_in=None)

  # Eval section
  config.evals = [
      ('minival', 'classification'),
      ('val', 'classification'),
      ('real', 'classification'),
      ('v2', 'classification'),
  ]
  eval_common = dict(
      pp_fn=pp_eval.format(lbl='label'),
      loss_name=config.loss,
      log_steps=1000,
  )

  config.minival = dict(**eval_common)
  config.minival.dataset = 'imagenet2012'
  config.minival.split = 'train[99%:]'
  config.minival.prefix = 'minival_'

  config.val = dict(**eval_common)
  config.val.dataset = 'imagenet2012'
  config.val.split = 'validation'
  config.val.prefix = 'val_'

  config.real = dict(**eval_common)
  config.real.dataset = 'imagenet2012_real'
  config.real.split = 'validation'
  config.real.pp_fn = pp_eval.format(lbl='real_label')
  config.real.prefix = 'real_'

  config.v2 = dict(**eval_common)
  config.v2.dataset = 'imagenet_v2'
  config.v2.split = 'test'
  config.v2.prefix = 'v2_'

  return config
\end{lstlisting}

\end{document}